\documentclass{article}
\usepackage{ijcai25}

\usepackage{times}
\usepackage{soul}
\usepackage{url}
\usepackage[hidelinks]{hyperref}
\usepackage[utf8]{inputenc}
\usepackage[small]{caption}
\usepackage{graphicx}
\usepackage{amsmath}
\usepackage{amsthm}
\usepackage{booktabs}
\usepackage{algorithm}
\usepackage{algorithmic}
\usepackage[switch]{lineno}
\usepackage{makecell}

\title{Consistency-Aware Padding for Incomplete Multi-Modal Alignment Clustering Based on Self-Repellent Greedy Anchor Search}

\author{
    Shubin Ma, Liang Zhao$^{*}$, Mingdong Lu, Yifan Guo, Bo Xu
    \affiliations
    School of Software Technology, Dalian University of Technology
    \emails
    1369830844@mail.dlut.edu.cn, liangzhao@dlut.edu.cn \\ mingdonglu@mail.dlut.edu.cn, qiaoan412@mail.dlut.edu.cn, BoXu@dlut.edu.cn
    \footnote{Corresponding Author}
}

\begin{document}

\maketitle

\begin{abstract}
    Multimodal representation is faithful and highly effective in describing real-world data samples' characteristics by describing their complementary information. However, the collected data often exhibits incomplete and misaligned characteristics due to factors such as inconsistent sensor frequencies and device malfunctions. Existing research has not effectively addressed the issue of filling missing data in scenarios where multiview data are both imbalanced and misaligned. Instead, it relies on class-level alignment of the available data. Thus, it results in some data samples not being well-matched, thereby affecting the quality of data fusion. In this paper, we propose the Consistency-Aware Padding for Incomplete Multimodal Alignment Clustering Based on Self-Repellent Greedy Anchor Search(CAPIMAC) to tackle the problem of filling imbalanced and misaligned data in multimodal datasets. Specifically, we propose a self-repellent greedy anchor search module(SRGASM), which employs a self-repellent random walk combined with a greedy algorithm to identify anchor points for re-representing incomplete and misaligned multimodal data. Subsequently, based on noise-contrastive learning, we design a consistency-aware padding module (CAPM) to effectively interpolate and align imbalanced and misaligned data, thereby improving the quality of multimodal data fusion. Experimental results demonstrate the superiority of our method over benchmark datasets. The code will be publicly released at https://github.com/Autism-mm/CAPIMAC.git.
\end{abstract}

\section{Introduction}
Multimodal clustering \cite{zhao2024learnable,zhou2024survey,wang2024scalable} \nocite{zhao2023deep}\nocite{zhao2023deep2} is an important research direction in machine learning and data mining, gaining significant attention due to its ability to leverage the features of multimodal data to enhance clustering results. To better describe the encoded features of each modality, multimodal clustering integrates each modality's complete information that is both aligned and complementary. However, a key challenge in multimodal clustering is addressing the incompleteness and misalignment \cite{tang2023multi,wang2023fualign} between modalities. Incompleteness refers to data being ignored or mislabeled as missing, often due to sensor failures, annotation errors, or preprocessing flaws. For example, In industrial monitoring, data may be missing due to sensor(infrared cameras and visual sensors) failures or environmental interference. Misalignment refers to the inability to establish a one-to-one correspondence between data items across different modalities, partially due to differing sensor collection frequencies or acquisition times. For example, in road traffic monitoring research \cite{dai2021temporal}, the system uses multiple images to determine whether a vehicle is violating regulations or speeding. However, the images of the same vehicle are not captured consecutively. Between the capture of two images, the camera continuously captures images of other vehicles. During data acquisition, incompleteness and misalignment occur simultaneously, significantly affecting subsequent multimodal data fusion\cite{li2023text,liu2024pro}. Current research focuses on how to align and fuse multimodal incomplete data with misalignment characteristics.

\begin{figure}
	\centering
		\includegraphics[scale=0.40]{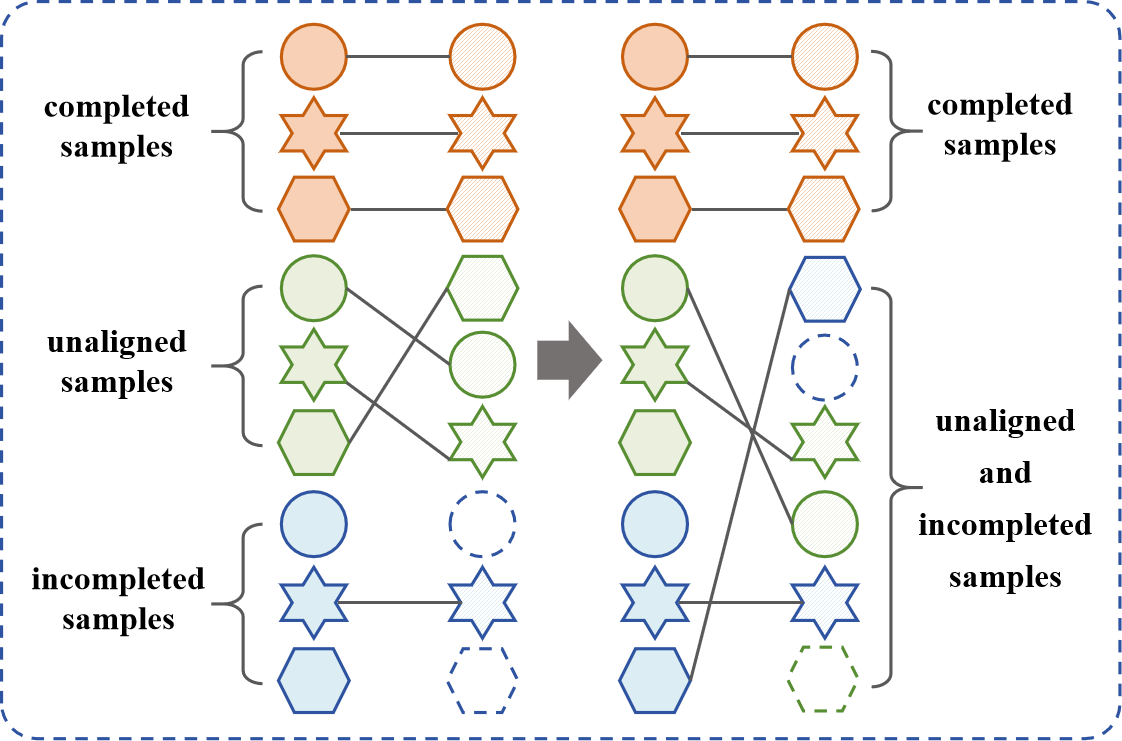}
	\caption{Incomplete partial alignment example graph.}
	\label{FIG:1}
\end{figure}

A toy example regarding incomplete and misaligned multimodal data clustering is shown in Figure \protect\ref{FIG:1},
and the same colour represents data of the same class, with solid lines indicating data matching. On the left, the matching of complete, misaligned, and incomplete multimodal data is shown; on the right, the matching of complete multimodal data and data with both incompleteness and misalignment is shown. From the right side of Figure \protect\ref{FIG:1}, it can be seen that in the incomplete and misaligned, the data order between the two modalities is inconsistent, and the data sizes are imbalanced.

Incompleteness and misalignment present significant challenges to multimodal clustering. To address these challenges, researchers have proposed various methods. In terms of incompleteness \cite{zong2021incomplete}\nocite{xu2022deep}\nocite{zhao2021intrinsic}\nocite{yang2024geometric}, Yang et al. \cite{yang2023cross} introduced one seminal work, CGMIMC, which effectively integrates structures across different modalities through consistency graph learning, thereby facilitating the completion of missing data.
Li et al. \cite{li2023incomplete} designed an IMvC method that simultaneously captures clustering information and modality-specific details through a dual-stream model and performs data recovery using prototypes from other modalities and data-prototype relationships in known modalities.
Huang et al. \cite{huang2024unified} restored missing data by dynamically learning feature similarity graphs and proposed a dynamic data quality assessment method, UNIFIER, based on semi-quadratic minimization to mitigate the impact of outliers and unreliable recovered data.
In terms of misalignment, Huang et al. \cite{huang2020partially} designed a PVC clustering model, introducing a differentiable alignment module that learns the optimal alignment strategy and can be integrated into deep learning models for alignment learning. Yang et al. \cite{yang2021partially} designed a new noise-robust contrastive loss model, MVCLN, which mitigates the impact of noise on data while enhancing the similarity between features of data from the same class using contrastive learning. Zeng et al. \cite{zeng2023semantic} designed the clustering model SMILE using information theory, which addresses completely misaligned multimodal data. By combining autoencoders with two cross-entropy losses, it reduces cross-modal differences while enhancing semantic distinction between different classes.
Zhao et al. \cite{zhao2024dynamic} proposed a progressive alignment learning model, DGPPVC, which, based on a Jaccard similarity variant and dynamic structural graph optimization, gradually identifies unknown correspondences between different modalities from simple to complex. He et al. \cite{he2024robust} proposed the VITAL model, which models each data sample as a Gaussian distribution in the latent space. By using variational inference and contrastive learning, the model preserves the commonality and specific semantics between modalities, enabling a comprehensive perception of the data.
To address both incomplete and misaligned multimodal data clustering, Yang et al. \cite{yang2022robust} designed a model called SURE. The model directly handles zeroed incomplete and misaligned multimodal data using Hungarian alignment and fills missing data by selecting and filling with the same data through KNN \cite{shi2018adaptive,li2023fast}. However, this method may lead to large similarity differences between aligned data pairs and lower-quality data pairs, affecting subsequent data fusion.

When addressing the misalignment of multimodal data, the number of data across modalities is consistent. However, multimodal incompleteness leads to an imbalance in the data across modalities, presenting a challenge for multimodal data alignment. To address the misalignment issue caused by data inconsistency and imbalances, enhance the coherence of data similarity—particularly when data missingness leads to misalignment between views, resulting in a sudden drop in similarity between data pairs (as illustrated in Figure 3)—and improve the quality of data fusion, we propose the Consistency-Aware Padding for Incomplete Multimodal Alignment Clustering Based on Self-Repellent Greedy Anchor Search, termed \textbf{CAPIMAC}. The proposed model primarily consists of three components: the self-repellent greedy anchor search module, the model training module, and the consistency-aware padding module. In the SRGASM, we select anchors \cite{qin2024dual,liu2024robust,li2023distribution}\nocite{yu2024dvsai,kang2020large} using a self-repellent random walk \cite{doshi2023self}, guided by a greedy algorithm. This method not only captures the complex local structures between data but also integrates the global information of the anchors. Then, we guide the learning of latent feature representations through a noise-contrastive loss \cite{zhao2021graph}\nocite{wang2022contrastive} function. Finally, in the CAPM, we introduce an innovative method by applying equal-weight Gaussian kernel interpolation to fill under-sampled modalities, preventing abrupt decreases in data-pair similarity and improving similarity coherence, thereby obtaining high-quality fused features. Experimental results demonstrate that the model effectively utilizes known data information for filling, thereby addressing the issues of incompleteness and misalignment in multimodal data.

The contributions of this paper can be summarized below.
\begin{itemize}
    \item We design the self-repellent greedy anchor search module to select anchors. This module selects anchors that are more structurally representative, reducing the complexity of subsequent computations. At the same time, the anchor selection process avoids local cycles and enhances the diversity of exploration.
    \item We introduce noise-contrastive loss to reduce the impact of false negative pairs, thereby guiding the learning of latent feature representations.
    \item We innovatively design the consistency-aware padding module to address the padding and alignment issue under imbalanced and misordered multimodal data. In this module, the equal-weight Gaussian kernel interpolation enhances the coherence of data-pair similarity, ensuring one-to-one alignment and fusion of the two modality sample data.
\end{itemize}
\section{Related Work}
\subsection{Random Walk}
Random walk is a fundamental statistical model that tracks random activity. The recursive formula for random walk in a multidimensional space is given by,
\begin{equation}
    \mathbf{p}(n) = P^n \mathbf{p}(0)
\end{equation}
$\mathbf{p}(n) = [ p_1(n), p_2(n)  \dots, p_N(n)] $ is the visit probability vector of the nodes at the \emph{nth} step, and $\mathbf{p}(0)$ is the initial state vector, representing the visit probabilities of the nodes at the \emph{0th} step. $P^n$ is the \emph{nth} power of the transition matrix, representing the system's state after \emph{n} steps of the random walk.

Random walk has various applications in machine learning. In adjacency graph construction, Wang et al. \cite{wang2020novel} reduced the time overhead in multi-label classification tasks by constructing a KNN-based random walk graph of training data. In combination with anchors, Liu et al. \cite{liu2010large} adopted an anchor-based random walk strategy, using anchors as "central" nodes for information propagation, optimizing the flow of information and improving learning efficiency. In node embedding features \nocite{dong2017metapath2vec,grover2016node2vec}, Aung et al. \cite{aung2024node} combined node embedding techniques with a random walk, enhancing the random walk process by using node embedding vectors. Additionally, random walk has wide applications in graph generation \cite{cai2022graph}, graph enhancement \cite{kim2023node}, anomaly detection \cite{liu2023fuzzy}, and graph sampling \cite{wang2021skywalker}.
\subsection{Gaussian Kernel Interpolation}

Gaussian kernel interpolation is commonly used for data smoothing, missing value imputation, or spatial data interpolation. The core idea is to estimate the value of a point using the weighted sum of Gaussian kernel functions. During Gaussian kernel interpolation, for a new interpolation point $x_{*}$ , the interpolated value $\hat{y}_{*}$ can be expressed as,
\begin{equation}
    \hat{y}_* = \sum_{i=1}^{n} w_i K(x_*, x_i)
\end{equation}
where $w_i$ is the weight, and $K(\cdot)$ denotes the Gaussian kernel function.

Gaussian kernel functions are widely used in multimodal clustering. For example, Liu et al. \cite{liu2020efficient} used Gaussian kernel functions to complete incomplete kernel matrices for learning consensus kernel matrices, addressing the issue of incomplete multimodal data. Ye et al. \cite{ye2017consensus}calculated the similarity between modalities based on the Gaussian kernel function and handled the nonlinear relationships in the data through the Gaussian kernel matrix, thereby increasing tolerance to missing data.
\section{The Proposed Method}
Given a dataset 
$\mathbf{X}=\{\mathbf{X}^{(1)},\mathbf{X}^{(2)},\mathbf{X}^{(3)},\dots,\mathbf{X}^{(v)}\}$ ,$\mathbf{X}^{(v)} \in \mathbf{R}^{n \times dv}$, represents the data of the \emph{vth} modality with \emph{n} samples and dimensionality 
\emph{dv}. To simulate the scenarios of misalignment and incompleteness, we introduce a shuffle matrix $\mathbf{U}=\{\mathbf{U}^{(1)},\mathbf{U}^{(2)},\mathbf{U}^{(3)},\dots,\mathbf{U}^{(v)}\}$ and a missing indicator matrix 
$\mathbf{C}=\{\mathbf{C}^{(1)},\mathbf{C}^{(2)},\mathbf{C}^{(3)},\dots,\mathbf{C}^{(v)}\}$ to process \textbf{X},where $\mathbf{U}^{(v)} \in \mathbf{R}^{n \times 1}$,$\mathbf{C}^{(v)} \in \mathbf{R}^{n \times 1}$.

To address the challenge of effectively filling missing values in incomplete and misaligned multimodal data, we propose the CAPIMAC framework, which consists of two key components:(1) the SRGASM, which selects critical data within each class through a self-repellent random walk strategy and utilizes a greedy algorithm to ensure that at least one key data is chosen from each class;(2) the CAPM, which employs an equal-weight Gaussian kernel padding strategy to ensure consistency in the data count across different modalities, thereby enhancing the coherence of data similarity.
\begin{figure*}
	\centering
		\includegraphics[scale=0.27]{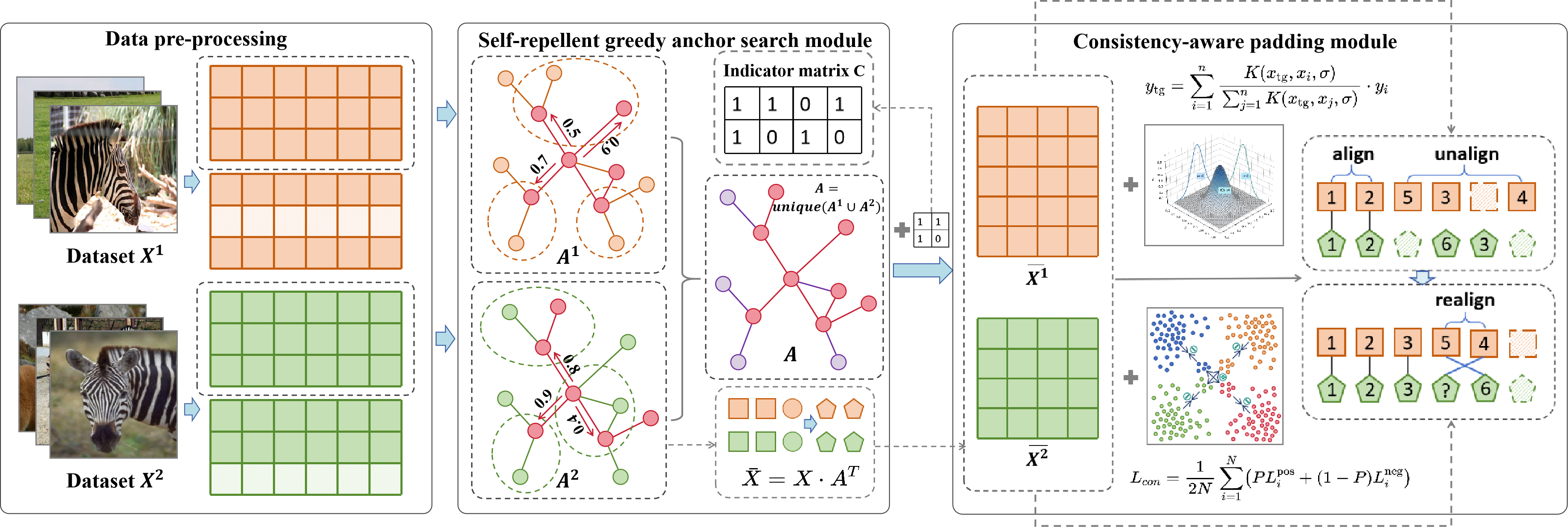}
	\caption{Incomplete and unaligned CAPIMAC model. In the Data Pre-processing, we handle incomplete and misaligned multimodal data. In the SRGASM, anchors are selected with a self-repelling random walk and greedy strategy, followed by data re-representation. In the CAPM, features learned via noise-contrastive loss are filled and aligned through Gaussian kernel interpolation.}
	\label{model}
\end{figure*}
\subsection{Self-Repellent Greedy Anchor Search Module}
In multimodal datasets, the high feature dimensionality of data leads to significant computational time during graph construction and training. To address this, we introduce anchors to simplify the representation of the original data. To better capture the complex local structures among the data and simultaneously identify the global optimal solution for the anchors, we propose the SRGASM. The module identifies anchors based on node density using a self-repellent random walk, with the number of steps and step size selected for each modality according to its characteristics,
\begin{equation}
(n_w, w_l) =
\begin{cases}
(20, 3) & \text{if } N < 100 \\
(10, 5) & \text{if } 100 \leq N < 1000 \\
(5, 10) & \text{if } 1000 \leq N < 10000 \\
(3, 20) & \text{if } N \geq 10000
\end{cases}
\end{equation}
where $n_w$ and $w_l$ denote the number of steps and step size, respectively, and \emph{N}
represents the number of data.
To describe the transition process between states, a state transition probability matrix is defined for each modality.
\begin{equation}
\begin{aligned}
    P_{ij}^{(v)} &= \cos(X_{ij}^{(v)}, X_{ij}^{(v)}) \\
    \text{s.t.} \quad &\sum_{j=1}^{n} P_{ij}^{(v)} = 1, \quad P_{ii}^{(v)} = 0 \quad (i = 1, 2, \dots, n)
\end{aligned}
\end{equation}
Specifically, \textbf{P} is a normalized probability matrix with zeros along the diagonal, and $cos(\cdot )$ represents the cosine similarity between data.

To avoid local cycles and enhance the diversity of exploration, we introduce a decay function $r_{ui}$ to reduce the probability of visiting previously visited nodes, where 
\begin{equation}
    r_{\mu_i}(x_i) = \left( \frac{x_i}{\mu_i} \right)^{-\alpha}, \quad \alpha \geq 0
\end{equation}
Here, ${u_i}$ represents the mean of the neighbouring data, and the parameter $\alpha \geq 0$
can be viewed as the strength of the self-repellent module in the self-repellent random walk transition kernel. When $\alpha>0$, 
$r_{ui}(x_i)$ decreases as $x_i$ increases, with larger $\alpha$ leading to faster decay. When $\alpha=0$, no decay is applied. At the same time, $r_{ui}(x_i)$ ensures that the probability matrix \textbf{P} exhibits scale invariance.

In the self-random walk sampling process, a smaller value of $\alpha$ improves the mixing property of the sampling, whereas a larger value of $\alpha$ enhances its efficiency. To strike a balance between the two, we set $\alpha=0.5$.Based on the transition probability matrix \textbf{P}, we obtain the visit matrix \textbf{V},
\begin{equation}
\begin{aligned}
    V &= \sum_{w=1}^{n_w} V_w ,V_w &= \sum_{t=1}^{w_l} P^t \times \pi
\end{aligned}
\end{equation}
Where \emph{w} represents the number of random walks, and \emph{t} denotes the number of steps in each random walk.

However, when all data in a particular class are significantly distant from data in other classes, it becomes challenging to select points from that isolated class during the self-repellent random walk, resulting in anchors that fail to effectively capture global information. To address this issue, we introduce a greedy algorithm in the self-repellent random walk process to expand the selection of anchors.

In the greedy algorithm, an anchor point $x_{i} \in X$ is first selected, and all nodes within a radius of 
\emph{d} are marked.s
\begin{equation}
    S_{k+1} = \{ x_i \} \cup \{ x_j \in X \mid d_{ij} \leq d \}
\end{equation}
Next, when selecting the next anchor point $x_{k} \in X$, it is required that its distance to all nodes in the previously selected anchor set \textbf{S} exceeds \emph{d}, meaning that a node outside the radius is chosen.
\begin{equation}
    S_{k+2} = S_{k+1} \cup \{ x_k \mid d_{kj} > d, \, \forall x_j \in S_{k+1} \}
\end{equation}
Here, $S_k$  denotes the set of all anchors selected at the \emph{kth} step. Once all nodes have been marked, the marks are reset, and the selection process is repeated until the required number of anchor points is reached.
\subsection{Model Training}
Since the anchors for each modality are selected independently, this results in mismatched anchor indices in the aligned portions. To standardize the representation dimensions across modalities, we unify the indices of the multiple modalities.
\begin{equation}
     I_u=unique(I^{(0)}\cup I^{(1)}\cup I^{(2)} \cdots I^{(v)} )
\end{equation}
$I_u$ denotes the deduplicated sum of the indices across multiple modalities.

To standardize the dimensional representation of multimodal data and facilitate improved training and graph construction, we re-represent the original data $\mathbf{X}^{(v)}$ using the selected consistency anchors $\mathbf{A}^{(v)}=\mathbf{X}^{(v)}[I_u]$,i.e.,
\begin{equation}
    \bar{X}=X\cdot A^T
\end{equation}
where $\mathbf{A}\in \mathbf{R}^{n_a\times dv}$, $n_a$ represents the number of anchors, which is equal to the size of $I_u$.

Due to the presence of noisy data in real-world multimodal datasets, we introduce noise-contrastive learning to mitigate the impact of noise. In the traditional contrastive loss,
\begin{equation}
    L_{con} = \frac{1}{n} \sum_{i=1}^{n} Y D_w^2 + (1 - Y) \max(m - D_w, 0)^2
\end{equation}
where, 
\begin{equation}
    \quad D_w(X_1, X_2) = \left( \sum_{i=1}^{n} (X_1^i - X_2^i)^2 \right)^{\frac{1}{2}}
\end{equation}
However, the traditional $L_{con}$ loss function has two main issues: first, Euclidean distance struggles to effectively measure similarity between high-dimensional data; and second, in the presence of noisy data, false negative pairs may occur, where two data points in a noisy pair actually belong to the same class. To address these issues, we propose an improved noise-contrastive\cite{yang2021partially} loss function,
\begin{align}
L_{recon} &= \frac{1}{2n} \sum_{i=1}^{n} \left( Y \cdot {D_c}^2 + (1 - Y) \cdot \frac{1}{m} \right. \\
&\quad \left. \max \left( a\cdot m \cdot {D_c} - {D_c}^3 , 0 \right)^2 \right) \notag
\end{align}
where \textbf{Y} represents the labels of positive and negative data pairs, $\mathbf{D_c}$ represents the cosine distance, and \emph{a} is a hyperparameter that controls the distance range within which false negative pairs can be mitigated.
\subsection{Consistency-Aware Padding Module}
\begin{figure}
	\centering
		\includegraphics[scale=0.55]{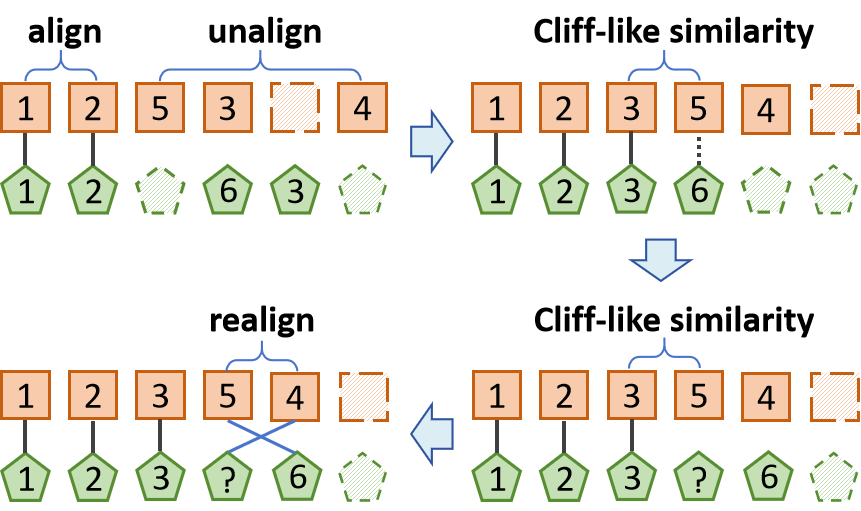}
	\caption{Flowchart of the Consistency-aware Padding Module: The process from 1 to 2 represents the sorting, from 2 to 3 represents the interpolation, and from 3 to 4 represents the realignment.}
	\label{kernel}
\end{figure}

In scenarios with incomplete and misaligned data, incompleteness leads to an imbalance in the number of multimodal data, while misalignment disrupts the order of data, making data padding challenging. To address these issues, we propose the CAPM, as shown in Figure \protect\ref{kernel}.

For $\mathbf{X}^{(v)}$, we sequentially apply misalignment and incompleteness processing to obtain the data $\mathbf{\hat{X}}^{(v)}$. Taking two modalities as an example,

\begin{equation}
    \hat{X}^{(v)}=\check{X}^{(v)}[C^{(v)}],\check{X}^{(v)}=X^{(v)}[U^{(v)}]
\end{equation}
Where, 
$\mathbf{U}$ is a shuffle index used to reorder the data in 
$\mathbf{X}$, and $\mathbf{C}$ is a \textbf{0},\textbf{1} index, where \textbf{0} indicates a missing data and \textbf{1} indicates an existing data, used to remove missing data. The resulting 
$\mathbf{X}^{(1)} \in \mathbf{R}^{nv_1 \times dv}$, $\mathbf{X}^{(2)} \in \mathbf{R}^{nv_2 \times dv}$,and 
$nv_1 \ne nv_2$.
Thus, the re-represented incomplete and misaligned data is,
\begin{equation}
    \bar{X}^{(v)}=\hat{X}^{(v)} \cdot {A^T}^{(v)}
\end{equation}
When the data is aligned, we need to compute a distance matrix \textbf{Z} for Hungarian alignment, where
\begin{equation}
\begin{aligned}
Z=1-{Z^{\prime}} ,
{Z^{\prime}} &= 
\begin{cases} 
\cos(\bar{X}^{(1)}, \bar{X}^{(2)}) & \text{if } nv_1 > nv_2 \\
\cos(\bar{X}^{(2)}, \bar{X}^{(1)}) & \text{if } nv_1 \le  nv_2
\end{cases}\\
\quad \text{s.t.} & \quad \sum_{j=1}^{n} {Z^{\prime}}_{ij} = 1
\end{aligned}
\end{equation}
For convenience, we use $n_s$ and $n_l$ to denote the smaller and larger data sizes, respectively.
\begin{equation}
    Z \in R^{n_s \times n_l},
    s.t.
\begin{cases} 
n_s=nv_2,n_l=nv_1 & \text{if } nv_1 > nv_2 \\
n_s=nv_1,n_l=nv_2 & \text{if } nv_1 \le  nv_2
\end{cases}\\
\end{equation}
We begin by using the Hungarian algorithm to find pairs of data with high similarity starting from the first row. However, due to class imbalance, some data do not form pairs with other data. Additionally, since missing data are randomly selected, there may be cases where data from different classes are aligned in the already aligned pairs, resulting in a significant disparity in similarity between data pairs. To maintain the coherence of data-pair similarity and reduce the occurrence of misaligned pairs from different classes, we rearrange the data-pair similarities, that is,
\begin{equation}
    Z^r=U^r \cdot Z
\end{equation}
Here, $ \mathbf{U^r}$ is a row transformation indicator matrix that rearranges the rows of \textbf{Z} in ascending order based on the data-pair distances. Next, we select $n_k$($n_k=n_l-n_s$) index pairs with the largest similarity differences and perform equal-weight Gaussian kernel interpolation between the indexed data.
The Gaussian kernel interpolation function is given by,
\begin{equation}
    y_{\text{target}} = \sum_{i=1}^{n_s} \frac{K(x_{\text{target}}, x_i, \sigma)}{\sum_{j=1}^{n_s} K(x_{\text{target}}, x_j, \sigma)} \cdot y_i
\end{equation}
where,
\begin{equation}
    K(x, x_i, \sigma) = \exp\left(-\frac{1}{2} \left( \frac{x - x_i}{\sigma} \right)^2\right)
\end{equation}
Through equal-weight interpolation, we obtain the new distance indicator matrix 
$\mathbf{\bar{Z^r}} \in \mathbf{R}^{n_l \times n_l}$. Subsequently, Hungarian alignment is performed based on ${\mathbf{\bar{Z^r}}}^T$
(Since no interpolation is performed on the modality with a length of $n_l$, virtual labels can be retrieved, which aids in performance evaluation.), followed by data fusion.

\section{Experiments}
In this section, we extensively evaluate the clustering property of the proposed method on five widely used multimodal benchmark datasets. The performance of CAPIMAC is compared with seven state-of-the-art multimodal methods (PVC\cite{huang2020partially}, MvCLN \cite{yang2021partially}, EGPVC \cite{zhao2023end}, SMILE \cite{zeng2023semantic}, SURE\cite{yang2022robust}, AMCP 
 \cite{zhao2024anchor}, DGPPVC \cite{zhao2024dynamic}) in terms of four clustering evaluation metrics.
\subsection{Experimental Settings}
Five datasets are used in our experiments, including 3Sources, yale\_mtv, BBCsports, Prokaryotic and 100leaves. In the experiments, we first perform data shuffling based on alignment rates, then apply missing data treatment to the shuffled portions. Since there are few models for handling incomplete and misaligned data, in the comparative experiments, we first apply missing data treatment to the models addressing misalignment, followed by Gaussian interpolation for filling and, finally, data shuffling.

The experiments were conducted in a Python 3.8 environment, using PyCharm 2022.2.3 as the integrated development environment.  The implementation was based on the deep learning framework PyTorch 1.8.2 with CUDA version 11.1. In the experiments, the alignment rate for all methods was set to 0.3, 0.5, and 0.7, while the incompleteness rate was set to 0.5. Since the primary challenge of multimodal incomplete and misaligned problems is alignment, with data filling being secondary, we used these settings to assess the model's performance under low, normal, and high alignment rates in the presence of typical data incompleteness. To validate the effectiveness of CAPIMAC in clustering, we conducted comparative and ablation experiments using ACC, NMI, ARI and F\_score(weighted) as performance metrics.

\subsection{Experimental Results and Analysis}
In this experiment, the best-optimized values for each metric are highlighted in bold, and the second-best values are underlined. BBC refers to the BBCsports dataset, Prok to the Prokaryotic dataset, and F1 to the F\_score (weighted).

\begin{table*}[ht]
    \renewcommand{\arraystretch}{0.95}
    \belowrulesep=0pt
    \aboverulesep=0pt
    \centering
    \scalebox{0.9}{
    \begin{tabular}{cc|cccccccc}
        \hline
            & \multicolumn{8}{c}{Methods} \\
        \hline
        Datasets &  & PVC 20' & MvCLN 21' & EGPVC 22' & SMILE 23'& SURE 23'& AMPC 24'&DGPPVC 24'& Ours \\
        \hline
            &ACC& 0.4686 & 0.3645 & 0.4278 & 0.3462 & 0.2823 & 0.3604 & \underline{0.4905}&\textbf{0.5168} \\
        3Sources&NMI&\underline{0.3354} & 0.1039 & 0.2182 & 0.0918 & 0.1256 & 0.1652 & 0.2270&\textbf{0.3707}\\
            &ARI& \underline{0.2239} & 0.0134 & 0.1586 & 0.0114 & 0.0019 & 0.0763 & 0.2202&\textbf{0.2538} \\
            &F1& \underline{0.4836} & 0.2809 & 0.4331 & 0.3091 & 0.3515 & 0.1692 & 0.4048&\textbf{0.5092} \\
    
        \hline
            &ACC & 0.4333 &0.3437 & \underline{0.4588} & 0.4109 & 0.3721 & 0.2655 & 0.2751 &\textbf{0.5229}\\
            yale\_mtv&NMI & 0.5182 & 0.4044 & \textbf{0.5335} & 0.4796 & 0.4529 & 0.3611 & 0.3749&\textbf{0.5445} \\
            &ARI & 0.2490 & 0.1185 & \underline{0.2656} & 0.1921 & 0.1564 & 0.0861 & 0.0976&\textbf{0.2735}\\
            &F1& 0.4189 & 0.3288 & \underline{0.4465} & 0.4090 & 0.3611 & 0.0458 & 0.2198&\textbf{0.5412} \\
    
        \hline
            
            &ACC& \underline{0.4475} & 0.3564 & 0.3575 & 0.3379 & 0.3060 & 0.4365 & 0.4248&\textbf{0.5119}\\
        BBC&NMI& 0.1789 & 0.0571 & 0.0697 & 0.0492 & 0.891 & \underline{0.2182} & 0.1139& \textbf{0.2871} \\
             &ARI&\underline{0.1245} & 0.0151 & 0.0366 & 0.0098 & 0.0152 & 0.1210 & 0.0745&\textbf{0.1744} \\
             &F1&\underline{0.4344} & 0.2593 & 0.3533 & 0.2852 & 0.3536 & 0.2275 & 0.3257&\textbf{0.531} \\

        \hline
            &ACC&0.3425 & \underline{0.4870} & 0.4011 & 0.4784 & 0.3428 & 0.3270 & 0.4182&\textbf{0.5823}\\
            Prok&NMI & 0.0308 & 0.1204 & 0.0784 & \underline{0.1618} & 0.0684 & 0.0110 & 0.0875&\textbf{0.2415} \\
            &ARI1 & 0.0144 & 0.0469 & 0.0419 & \underline{0.0731} & 0.0230 & 0.0042 & 0.0651&\textbf{0.2059} \\
             &F1& 0.3646 & 0.4541 & 0.4191 & 0.4788 & \underline{0.5252} & 0.2793 & 0.4520&\textbf{0.5838} \\
        \hline
            &ACC & 0.2329 & 0.2948 & 0.2958 & 0.3535 & 0.3195 & 0.2634 & \underline{0.4306}&\textbf{0.5345}\\
       100leaves &NMI & 0.5492 & 0.5750 & 0.6143 & 0.6605 & 0.1676 & 0.6320 & \underline{0.6871} & \textbf{0.7705}\\
            &ARI &0.1034 & 0.1302 & 0.1613 & 0.2143 & 0.0587 & 0.1557 &\underline{0.2824}&\textbf{0.3827} \\
             &F1& 0.2203 & 0.2977 & 0.2832 & 0.3401 & 0.3876 & 0.0092 &\underline{0.4171}&\textbf{0.5295} 
            \\
        \hline
    \end{tabular} 
    }
    \caption{The clustering performance(ACC, NMI, ARI and F1) of various algorithms on different datasets. The alignment rate is 0.5; the incomplete rate is 0.5.}
    \label{tbl0.5}
\end{table*}
Tables \protect\ref{tbl0.5},\protect\ref{tbl:0.3},\protect\ref{tbl:0.7} report the experimental results of eight different methods on five multimodal datasets, evaluated using metrics such as ACC, NMI, ARI and F1. These metrics assess the methods' performance in clustering tasks. From these results, we obtain the following conclusions.
\begin{itemize}
    \item Tables \protect\ref{tbl0.5},\protect\ref{tbl:0.3},\protect\ref{tbl:0.7} present the clustering results of eight models on five datasets. Among the four metrics evaluated on these datasets, our model achieved the best performance in the majority of the results, demonstrating its effectiveness in addressing the multimodal incomplete and misaligned data problem.
    \item  As shown in Tables \protect\ref{tbl0.5},\protect\ref{tbl:0.3},\protect\ref{tbl:0.7}, our model achieves excellent results on the 100leaves dataset under incomplete data conditions with low, medium, and high alignment rates. This demonstrates that our model can maintain strong clustering performance even when dealing with datasets with a large number of classes.
    \item When addressing the multimodal misalignment and incompleteness problem, the clustering results of our model show a slight decrease as the alignment rate decreases, yet it still achieves the best performance in the majority of clustering results. This demonstrates that our model has strong robustness and can handle incomplete alignment fusion problems under varying alignment rates.
    \item Since our model requires selecting anchors from complete data for data re-representation, it is unable to handle incomplete and misaligned datasets with excessively low alignment rates. This is why the model's performance shows a slight decline when the alignment rate is 0.3.
    \item The experimental results demonstrate that the selected anchors effectively capture both the local and global information of the modality data. Additionally, equal-weight Gaussian kernel interpolation addressed the issue of incoherent similarity, further improving clustering performance.
\end{itemize}
\begin{table}[!ht]
\small
\centering
\renewcommand{\arraystretch}{0.95}
\tabcolsep=0.08cm

\begin{tabular}{l c c c c c c} 
 \hline
 Method & & 3Sources & yale\_mtv & BBC & Prok &100leaves \\ 
 \hline
 PVC  & ACC & 0.3710 & 0.4364 & 0.4096 & 0.3998&0.1848\\ 
      & NMI & 0.2110 & 0.5036 & 0.1376 & 0.0613&0.4875 \\ 
      & ARI & 0.1096 & 0.2212 & 0.0893 & 0.0522&0.0521 \\ 
      & F1 &  0.3757& \underline{0.4339} & 0.3869 & 0.4328&0.1812 \\ 
MvCLN & ACC & 0.3538 & 0.3297 & 0.3638 & \underline{0.5015}&0.2964 \\ 
      & NMI & 0.0996 & 0.4106 & 0.0610 & 0.1446&0.5832 \\ 
      & ARI & 0.0081 & 0.1278 & 0.0234 & 0.0577&0.1351 \\ 
      & F1 &  0.2066 & 0.3080 & 0.2838 & 0.4671&0.2988 \\ 
EGPVC & ACC & 0.4414 & \underline{0.4424} & 0.3645 & 0.4726&0.3025  \\ 
      & NMI & 0.2247 & \underline{0.5228} & 0.0696 & \underline{0.1740}&0.6054 \\ 
      & ARI & 0.1373 & \textbf{0.2539} & 0.0320 & 0.1166&0.1574 \\ 
      & F1 &  \underline{0.4286} & 0.4270 & 0.3238 & 0.4766&0.2891 \\ 
SMILE & ACC & 0.3643 & 0.4388 & 0.4004 & 0.4885&0.3487\\ 
      & NMI & 0.1122 & 0.4995 & 0.1211 & 0.1555&\underline{0.6632}\\ 
      & ARI &0.0016 & 0.2183 & 0.0980 & \underline{0.1213}&0.2104\\ 
      & F1 &  0.3222 & 0.4336& 0.3383 & 0.4970&0.3421 \\ 
SURE  & ACC & 0.3426 & 0.3636 & 0.3596 & 0.3733&0.3092 \\ 
      & NMI & 0.1834 & 0.4361& 0.1558 & 0.0866&0.1539\\ 
      & ARI  &0.0417& 0.1768 & 0.0831 & 0.0415&0.0351\\ 
      & F1 & 0.3971& 0.4103 & \underline{0.3982} & \underline{0.4986}&0.3417\\ 
AMPC  & ACC &0.3266& 0.2497 & \underline{0.4674} & 0.3323&0.2019 \\ 
      & NMI &0.1418& 0.3302 & \textbf{0.2767} & 0.0112&0.5370\\ 
      & ARI &0.0426& 0.0580& \textbf{0.1753} & 0.0016&0.0763\\ 
      & F1 & 0.1662& 0.0640 & 0.2323 & 0.2871&0.0058 \\ 

DGPPVC& ACC &\underline{0.4651}& 0.1830 & 0.3840 & 0.4145&\underline{0.3838} \\ 
      & NMI &\underline{0.2386}& 0.2480 & 0.0714 & 0.0748&0.6607 \\ 
      & ARI  &\underline{0.1706}& 0.0205 & 0.0389 & 0.0551&\underline{0.2404}\\ 
      & F1 & 0.3747& 0.1640 & 0.2555 & 0.4413&\underline{0.3677}\\ 

Ours & ACC &\textbf{0.5299}& \textbf{0.4830} & \textbf{0.4953} & \textbf{0.5979}&\textbf{0.4887} \\ 
      & NMI &\textbf{0.3926}& \textbf{0.5277} & \underline{0.2399} & \textbf{0.2755}&\textbf{0.7487}\\ 
      & ARI  &\textbf{0.2720}& \underline{0.2358} & \underline{0.1712} & \textbf{0.2637}&\textbf{0.3426}\\ 
      & F1 &\textbf{0.5416}& \textbf{0.5068} & \textbf{0.5066} & \textbf{0.6068}&\textbf{0.4847}\\ 
 \hline
\end{tabular}
\caption{The clustering performance(ACC, NMI, ARI and F1) of various algorithms on different datasets. The alignment rate is 0.3; the incomplete rate is 0.5. }
\label{tbl:0.3}
\end{table}

\begin{table*}[!ht]
    \renewcommand{\arraystretch}{0.95}
    \belowrulesep=0pt
    \tabcolsep=0.4cm
    \aboverulesep=0pt
    \centering
    \scalebox{0.9}{
    \begin{tabular}{cc|cccccccc}
         \hline
        \multicolumn{2}{c} {Metric}& \multicolumn{2}{c}{ACC} & \multicolumn{2}{c}{NMI} & \multicolumn{2}{c}{ARI} & \multicolumn{2}{c}{F1} \\
        \hline
         Rate & Datasets & No IPT & IPT & No IPT & IPT & No IPT & IPT & No IPT & IPT \\
       
        \hline
         & 3sources & 0.4745 & \textbf{0.5299} & 0.3201 & \textbf{0.3926} & 0.2344 & \textbf{0.2720} & 0.4967 & \textbf{0.5416} \\
        & BBC & 0.4554 & \textbf{0.5073} & 0.1768 & \textbf{0.2485} & 0.1246 & \textbf{0.1827} & 0.4604 & \textbf{0.5179} \\
        0.3align& yale\_mtv & 0.4741 & \textbf{0.4830} & 0.5144 & \textbf{0.5277} & 0.2064 & \textbf{0.2358} & 0.4966 & \textbf{0.5068} \\
        & Prokaryotic & 0.4703 & \textbf{0.6101} & 0.0517 & \textbf{0.2842} & 0.0748 & \textbf{0.2860} & 0.4722 & \textbf{0.6201} \\
        & 100leaves & 0.3942 & \textbf{0.4887} & 0.6771 & \textbf{0.7487} & 0.2272 & \textbf{0.3426} & 0.3869 & \textbf{0.4847} \\
        \hline
        & 3sources & 0.4897 & \textbf{0.5168} & 0.3078 & \textbf{0.3707} & 0.2396 & \textbf{0.2538} & 0.5092 & \textbf{0.5273}  \\
        & BBC & 0.4527 & \textbf{0.4722} & 0.1393 & \textbf{0.2406} & 0.1246 & \textbf{0.1597} & 0.4701 & \textbf{0.4834} \\
         0.5align& yale\_mtv & 0.4786 & \textbf{0.5229} & 0.5074 & \textbf{0.5445} & 0.2085 & \textbf{0.2735} & 0.4917 & \textbf{0.5427} \\
        & Prokaryotic & 0.4937 & \textbf{0.5823} & 0.0648 & \textbf{0.2415} & 0.0186 & \textbf{0.2059} & 0.4191 & \textbf{0.5838} \\
        & 100leaves & 0.3977 & \textbf{0.5345} & 0.6771 & \textbf{0.7705} & 0.2347 & \textbf{0.3827} & 0.3951 & \textbf{0.5295} \\
        \hline
         & 3sources & 0.4808 & \textbf{0.5531} & 0.4279&\textbf{0.4330}  & 0.2598 & \textbf{0.3209} & 0.4982 & \textbf{0.5719} \\
        & BBC & 0.4864 & \textbf{0.5027} & 0.2173 & \textbf{0.2768} & 0.1825 & \textbf{0.1901} & 0.4861 & \textbf{0.5222} \\
        0.7align& yale\_mtv & 0.5033 & \textbf{0.5411} & 0.5392 & \textbf{0.5614} & 0.2623 & \textbf{0.3033} & 0.5254 & \textbf{0.5629} \\
        & Prokaryotic & 0.5520 & \textbf{0.6259} & 0.1035 & \textbf{0.2976} & 0.1290 & \textbf{0.2930} & 0.5374 & \textbf{0.6386} \\
        & 100leaves & 0.3744 & \textbf{0.5749} & 0.6703 & \textbf{0.7959} & 0.2210 & \textbf{0.4424} & 0.3716 & \textbf{0.5648} \\
        \hline
    \end{tabular}
    }
    \caption{Ablation Experiment of CAPIMAC on Whether to Use the Consistency-Aware Padding Module at Alignment Rates of 0.3, 0.5, and 0.7.}
    \label{tbl_abl}
\end{table*}

\begin{table}[!ht]
\small
\centering
\renewcommand{\arraystretch}{1}
\tabcolsep=0.08cm

\begin{tabular}{l c c c c c c} 
 \hline
  Method & & 3Sources & yale\_mtv & BBC & Prok &100leaves \\ 
 \hline
 PVC  & ACC & 0.4414 & \underline{0.4497} & \underline{0.4113} & 0.3386&0.2504\\ 
      & NMI & \underline{0.3303} & \underline{0.5287} & \underline{0.1775} & 0.0442&0.5736 \\ 
      & ARI & \underline{0.2053} & \underline{0.2607} & \underline{0.1021} & 0.0162&0.1227 \\ 
      & F1 &  \underline{0.4603} & \underline{0.4394} & 0.3772 & 0.3630&0.2282 \\ 
MvCLN & ACC & 0.3539 & 0.3224 & 0.3521 & \underline{0.5158}&0.2971 \\ 
      & NMI & 0.0874 & 0.3860 & 0.0495 & 0.0986&0.5715 \\ 
      & ARI & 0.0069 & 0.1034 & 0.0082 & 0.0343&0.1319 \\ 
      & F1  & 0.2530 & 0.3101 & 0.2445 & 0.4629&0.3015 \\ 
EGPVC & ACC & 0.4201 & 0.4412 & 0.3862 & 0.4688&0.3296  \\ 
      & NMI & 0.2183 & 0.5170 & 0.0994 & 0.1057&0.6261 \\ 
      & ARI & 0.1380 & 0.2484 & 0.0539 & 0.0535&0.1811 \\ 
      & F1 &  0.4229 & 0.4244 & \underline{0.3784} & 0.4526&0.3205  \\ 
SMILE & ACC & 0.3118 & 0.4115 & 0.3521 & 0.4554&0.3800 \\ 
      & NMI & 0.0863 & 0.4768 & 0.0595 & \underline{0.1598}&0.6812 \\ 
      & ARI  &0.0126 & 0.1974 & 0.0165 & 0.0567&0.2395 \\ 
      & F1 & 0.2896 & 0.4032 & 0.2960 & 0.4515&0.3725  \\ 
SURE  & ACC &0.2858 & 0.3951 & 0.3177 & 0.3434&0.3240  \\ 
      & NMI &0.1134 & 0.4712 & 0.0856 & 0.0796&0.1627 \\ 
      & ARI &0.0152 & 0.1848 & 0.0160 & 0.0375&0.0333 \\ 
      & F1  &0.3805 & 0.4085 & 0.3674 & \underline{0.5363}&0.3935 \\ 
AMPC& ACC   &0.3053 & 0.2976 & 0.3574 & 0.3438&0.2250  \\ 
      & NMI &0.1099 & 0.4159 & 0.1123 & 0.0079&0.5725 \\ 
      & ARI &0.0376 & 0.1210 & 0.0486 & 0.0028&0.1049 \\ 
      & F1  &0.1842 & 0.0715 & 0.1955 & 0.2968&0.0108  \\ 

DGPPVC& ACC & \underline{0.4716} & 0.2497 & 0.3874 & 0.3924&\underline{0.4341}  \\ 
      & NMI & 0.2445 & 0. 3825& 0.0884 & 0.0735&\underline{0.6982}  \\ 
      & ARI & 0.1915 & 0.0903 & 0.0394 & \underline{0.0599}&\underline{0.3030} \\ 
      & F1  & 0.3794 & 0.2030 & 0.2607 & 0.4293&\underline{0.4157} \\ 

Ours&ACC& \textbf{0.5531} & \textbf{0.5411} & \textbf{0.5027} &\textbf{ 0.6259}&\textbf{0.5749}  \\ 
      & NMI &\textbf{0.4279} & \textbf{0.5614} &\textbf{ 0.2768} & \textbf{0.2976}&\textbf{0.7959 }\\ 
      & ARI & \textbf{0.3209} & \textbf{0.3033} & \textbf{0.1901} & \textbf{0.2930}&\textbf{0.4424} \\ 
      & F1  &\textbf{0.5719} &\textbf{ 0.5629} & \textbf{0.5222} & \textbf{0.6386}&\textbf{0.5648} \\ 
 \hline
\end{tabular}
\caption{The clustering performance(ACC, NMI, ARI and F1) of various algorithms on different datasets. The alignment rate is 0.7; the incomplete rate is 0.5. }
\label{tbl:0.7}
\end{table}
\subsection{Time Complexity Analysis}
After obtaining the similarity graph, KNN needs to find the \emph{k} nearest neighbors for each data point $x_i$. The \emph{k} smallest distances are identified through sorting, which has a complexity of $O(nlogn)$, resulting in an overall complexity of 
$O(n^2logn)$. However, random walk typically explores only within the local neighbourhood, with a complexity of $O(nlogn)$ or $O(n\cdot k)$.
\subsection{Ablation Experiment}
Table \protect\ref{tbl_abl} presents the ablation experiment on whether the model incorporates the CAPM.
No IPT indicates that missing data is discarded directly without applying equal-weight Gaussian kernel interpolation; IPT indicates that equal-weight Gaussian kernel interpolation was applied.

As shown in Table \protect\ref{tbl_abl}, the CAPM has improved clustering performance to some extent under low, medium, and high alignment rates. Essentially, the CAPM not only utilizes all available data information, increasing data utilization, but also addresses the issue of sharp similarity gaps in imbalanced data, ultimately enhancing the quality of multimodal data fusion.
\section{Conclusion}

In this paper, we propose CAPIMAC to address the feature fusion problem under the conditions of incomplete and misaligned multimodal data. By designing the SRGASM, we select anchors that capture both local and global information from the modalities. We resolve the alignment and padding issue under misaligned and imbalanced multimodal data through the CAPM, thus improving the quality of the fused features. Our method provides a new perspective for solving the filling problem of misordered and imbalanced data. In the future, we plan to extend CAPIMAC to explore the interpolation and alignment issues in cases where multimodal data are completely misaligned and partially incomplete.

\appendix

\section*{Acknowledgments}
This work is supported by the Science and Technology Project of Liaoning Province (2024JH2/102600027, 2023JH2/101700363)
and the Science and Technology Project of Dalian City
(2024JJ12GX025, 2023JJ12SN029 and 2023JJ11CG005).
\bibliographystyle{named}
\bibliography{ijcai25}

\end{document}